\documentclass[10pt,twocolumn,letterpaper]{article}

\usepackage{iccv}
\usepackage{times}
\usepackage{epsfig}
\usepackage{graphicx}
\usepackage{amsmath}
\usepackage{amssymb}
\usepackage{utfsym}

\usepackage{bm}
\usepackage{makecell}

\usepackage{algorithm, algorithmic}
\usepackage[pagebackref=true,breaklinks=true,letterpaper=true,colorlinks,bookmarks=false]{hyperref}
\ificcvfinal\fi

\begin{document}

%%%%%%%%% TITLE
\title{ADMap: Anti-disturbance framework for vectorized HD map construction}

\author{Haotian Hu\\
Leapmotor\\
{\tt\small hu\_haotian@leapmotor.com}
\and
Fanyi Wang\\
Zhejiang university\\
{\tt\small 11730038@zju.edu.cn}
\and
Yaonong Wang\\
 Leapmotor\\
{\tt\small wang\_yaonong@leapmotor.com}
\and
Laifeng Hu\\
Leapmotor\\
{\tt\small hu\_laifeng@leapmotor.com}
\and
Jingwei Xu\\
Leapmotor\\
{\tt\small xu\_jingwei@leapmotor.com}
\and
Zhiwang Zhang\\
NingboTech University\\
{\tt\small zhiwang.zhang@nit.zju.edu.cn}
}

\maketitle
\ificcvfinal\thispagestyle{empty}\fi

% 在自动驾驶领域，在线高精地图重建对于规划和预测任务具有重要意义，近期的工作构建了许多高性能的高精地图重建模型来满足这一需求。然而实例矢量内部的点序由于预测偏差可能会出现抖动或锯齿现象，从而影响后续任务。因此，本文提出了 Anti-disturbance map framework(ADMap)。该框架包括了三个用于缓解点序抖动的模块：Multi-Scale Perception Neck/Instance Interactive Attention(IIA)和Vector Direction Difference Loss(VDDL)。通过级联的方式探索实例间和实例内部的点序关系，模型更好地监督了点序的预测过程。我们在nuScenes和Argoverse2中验证了ADMap的卓越性能，。。。

\begin{abstract}
  In the field of autonomous driving, online High-definition (HD) map construction is crucial for planning tasks. Recent research has developed several high-performance HD map construction models to meet this necessity. However, the point sequences within the instance vectors may be jittery or jagged due to prediction bias, which can impact subsequent tasks. Therefore, this paper proposes the Anti-Disturbance Map construction framework (ADMap). To mitigate point sequence jitter, the framework consists of three modules: Multi-scale Perception Neck (MPN), Instance Interactive Attention (IIA), and Vector Direction Difference Loss (VDDL). By exploring the point sequence relationships between and within instances in a cascading manner, the model can monitor the point sequence prediction process more effectively. ADMap achieves state-of-the-art performance on the nuScenes and Argoverse2 datasets. Extensive results demonstrate its ability to produce stable and reliable map elements in complex and changing driving scenarios.
\end{abstract}

\section{Introduction}
%在自动驾驶领域，矢量化高精地图重建是一项关键任务，其将传感器收集的信息重建为实例级矢量表示（车道线/路边沿和人行横道），使得车辆在行驶过程中能够捕捉到详细的道路拓扑结构和地面语义信息，并有效运用于下游预测和规控任务。
%近年来，高清地图得到了长足发展，早期工作【】预测了稠密的地面信息，这导致模型计算和标注的冗余。HDMapNet[]将稠密的像素分割结果分组为稀疏矢量化实例，但这需要复杂的后处理过程。VectorMapNet首次预测矢量化实例，利用自回归解码器有序的预测实例点。MapTR端到端的预测矢量化实例并有效解决了点序方向不同所造成的特征模糊。MapTRv2【】。。。

In the field of autonomous driving, high-definition (HD) map construction ~\cite{li2021hdmapnet, liu2022vectormapnet, maptrv2, xie2023mvmap, yu2023ScalableMap} is a crucial task. This process involves converting the sensor-collected information into instance-level vector representations, such as lane lines, road boundaries, and pedestrian crossings. These representations enable the vehicle to capture detailed road topology and ground semantics information while driving, which can be effectively applied to downstream regulation tasks.
In recent years, early HD maps construction works~\cite{prework1, prework2, prework3, prework4, pre5, pre6} predicted dense ground information, which resulted in redundancy in model computation and annotation. HDMapNet~\cite{li2021hdmapnet} groups dense pixel segmentation results into sparse vectorized instances, but this requires a complex post-processing process. VectorMapNet~\cite{liu2022vectormapnet} predicts vectorized instances for the first time, using an autoregressive decoder to predict the instance points in an ordered manner. MapTR~\cite{MapTR} predicts vectorised instances end-to-end and resolves feature ambiguities caused by different point order directions effectively.  MapTRv2~\cite{maptrv2} adds decoupled self-attention to capture intra-instance point relations in parallel.

\begin{figure}[tb]
\centering
\includegraphics[width=\linewidth]{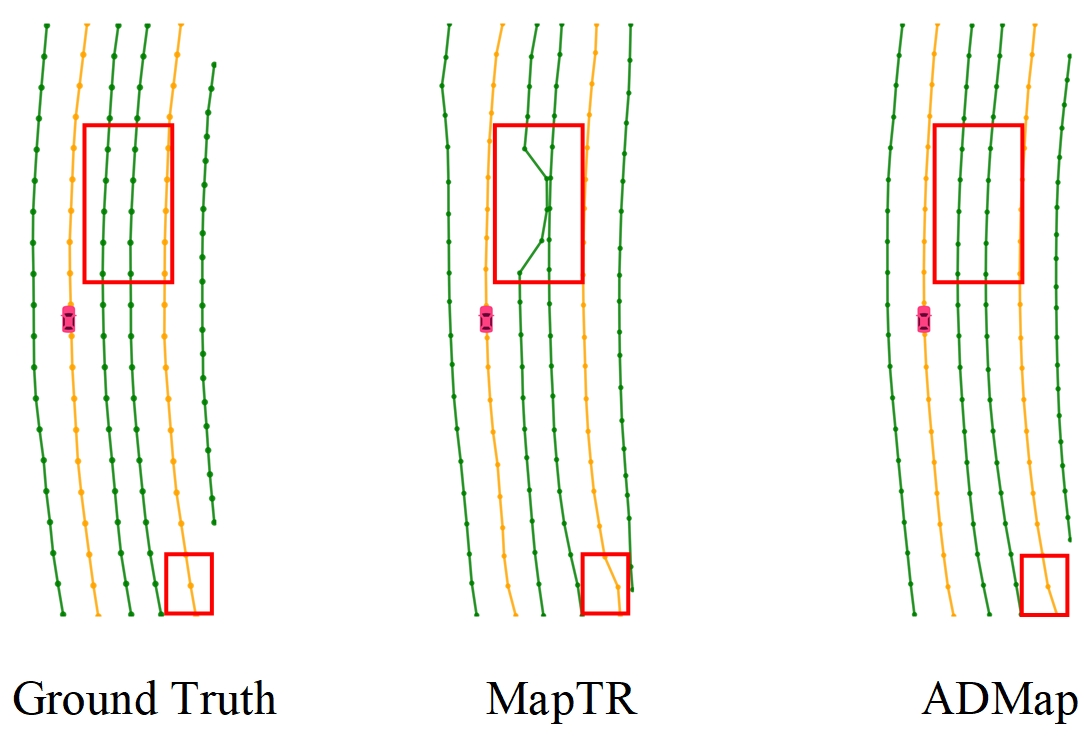}
   \caption{Performance and visualization comparison between ADMap and baseline. The two endpoints of the line segment in the left figure indicate the results of the multi-modal and camera-only frameworks, respectively. The right figure shows that ADMap effectively mitigates the point sequence jitter problem.} % ADMap和baseline间的可视化对比，ADMap有效的缓解了点序抖动问题。
   \label{fig:introduction}
\end{figure}

%如图1所示，我们观察到，实例中的预测点往往会不可避免的出现抖动或偏移现象，这种抖动会导致重建后的实例矢量变得扭曲或锯齿状，严重影响了在线高精地图的质量和实用性。我们认为，其原因在于现有模型并未充分考虑实例间和实例内部的交互方式，实例点和地图拓扑信息的不完全交互导致了预测点的不准确。此外，仅通过L1 loss这类距离监督无法有效的利用点序间的几何关系来约束点序的预测过程，网络需要利用各点间的矢量线段的特征来更精细的约束点序位置。
%先前的特征提取方法感受野有限
As shown in Figure~\ref{fig:introduction}, we observe that the predicted points in the instances tend to be inevitably jittered or shifted, causing the constructed instance vectors to become distorted or jagged, which seriously affects the quality and utility of the online high-precision maps. It is suggested that existing models may not fully consider inter-instance and intra-instance interactions, resulting in incomplete interaction between instance points and map topology information, leading to inaccurate prediction points. Furthermore, using L1 loss\cite{Wang2024GeMap, Gao2024Complementing, MapTR} or cosine embedding loss\cite{MapTR, maptrv2} alone for distance supervision does not effectively utilize the geometric relationship between point sequences to constrain the prediction process. To address this, the network must utilize the characteristics of the vector line segments between points to more finely constrain the position of the point sequences.

%为了缓解这一问题，我们创新的提出了Anti-disturbance map framework(ADMap)，该框架通过multi-scale perception neck（MPN）/instance interactive attention(IIA)和vector direction difference loss(VDDL)来更精细的预测点序拓扑结构。MPN在不增加推理时间的前提下，是网络更好的捕捉BEV map中包含的多尺度特征以更准确的重建场景中尺寸差异较大的不同实例。IIA灵活编码了实例级和点级别信息，实例嵌入间的特征交互进一步帮助网络捕捉点级嵌入间的关注，更准确的点级信息使得重建更准确。VDDL建模了实例点和矢量方向差之间的关联，并使用矢量方向差作为损失来更精细地约束点序的重建过程。此外，真实矢量方向差被用于赋予实例中各点不同权重，以保证模型可以更好的捕捉到场景中剧烈变化的地图信息。
To address this issue, we propose the Anti-Disturbance Map construction (ADMap) framework. ADMap framework consists of Multi-scale Perception Neck (MPN), Instance Interactive Attention (IIA) and Vector Direction Difference Loss (VDDL) to more accurately predict point sequence topology. 

MPN enables the network to capture multi-scale features in the BEV map, improving the accuracy of constructing instances with significant size differences in the scene without increasing inference time. IIA flexibly encodes instance-level and point-level information, feature interactions between instance embeddings further help the network capture the relationships between point-level embeddings, and more accurate point-level information makes the construction more accurate. VDDL models the association between instance points and vector direction differences, and uses vector direction differences as losses to constrain the construction process of point sequences more precisely. Additionally, the difference in real vector direction is utilised to assign varying weights to the points in the instances, ensuring that the model can more effectively capture the rapidly changing map information in the scene.

%我们在nuScenes和Argoverse2中验证了ADMap的有效性。与现有的重建矢量化高精地图模型相比，ADMap在nuScenes和Argoverse2中都实现了领先的性能。在nuScenes中，ADMap在camera-only和多模态框架下相较于baseline分别提高了4.2\%和5.5\，ADMapv2不仅减少了推理延迟，还有效提高了baseline性能。在Argoverse2中，ADMap也有出色表现。ADMapv2的mAP提升为()\%，同时FPS还保持了()。这证明了ADMap是一个高效且高精度的框架，其可以在复杂的场景中生成精确且顺滑的地图拓扑。
We validate the effectiveness of our proposed ADMap in nuScenes benchmark \cite{nuscenes2019} and Argoverse2 benchmark \cite{Argoverse2}.
ADMap achieves leading performance in both nuScenes and Argoverse2 compared to existing constructed vectorized HD map models. 
In nuScenes , ADMap improved performance by 4.2\% and 5.5\% in camera-only and multimodal frameworks, respectively, compared to the baseline method MapTR. ADMapv2 not only reduced inference latency but also improved performance of baseline method MapTRv2. ADMap also performed well in Argoverse2. ADMapv2 improves mAP by 62.9\% while FPS maintaining 14.8, demonstrating that ADMap is an efficient and high-precision framework for generating accurate and smooth map topology in complex scenes.

% 1.提出了端到端的ADMap，其重建了更稳定的矢量化高精地图。
% 2.MPN在不增加推理资源的情况下更精细的捕捉多尺度信息，IIA完成了实例间和实力内部信息的有效交互，有效缓解实例点位置偏移问题提。VDDL建模了矢量角度差，并利用拓扑信息监督点序位置的重建过程。
% 3.ADMap实现了矢量化高精地图的实时重建，并且在nuScenes基准和Argoverse2基准中达到了最佳性能。
The contributions of this paper are summarized as follows:
\begin{itemize}
    \item End-to-end framework ADMap is proposed, which constructs more stable vectorized HD maps .
    
    \item MPN captures multi-scale information more precisely without increasing computational resources, IIA achieves effective interaction between inter-instance and intra-instance information to alleviate the problem of instance point position offset. VDDL models the vector direction difference and supervises theconstruction process of point order position using topological information.
    
    \item ADMap enables real-time construction of vectorized HD maps and achieves state-of-the-art in both the nuScenes and Argoverse2 benchmarks.
\end{itemize}

\section{Related Work}
% 车道线检测
\subsection{Lane detection}\label{lane_detection}
% 在先前的工作中，车道线检测往往被视为一个独立的任务，其通过摄像头,激光雷达等传感器获取信息，并对车道线进行识别和定位。LaneNet【】提出用于2D车道线的一种车道线语义分割并聚类的方法。3D-LaneNet【】是单目3D车道线领域的开创性工作，其提出一种新型的双路径结构，在网络内部实现特征的Inverse Perspective Mapping(IPM)投影。GenLaneNet【】优化了3D-LaneNet的anchor representation，在更可靠的坐标系中使用anchor预测3D Lane。PerFormer【】提出了统一的二维和三位车道检测框架，并将Transformer引入空间变换模块中，以获得更鲁棒的特征。
In previous research, lane line detection was typically considered a standalone task. Information was gathered through sensors such as cameras and lidar to identify and locate lane lines. LaneNet \cite{neven2018lanenet} proposes a semantic segmentation of 2D lane lines and clusters them. 3D-LaneNet \cite{3dlanenet} is a pioneering work in the field of monocular 3D lane lines, which proposes a new type of dual-path structure that implements inverse perspective mapping (IPM) projection of features inside the network. GenLaneNet \cite{guo2020genlanenet} optimizes the anchor representation of 3D-LaneNet and uses the anchor to predict 3D Lane in a more reliable coordinate system. PersFormer \cite{chen2022persformer} proposes a unified framework for 2D and 3D lane detection. The framework introduces transformer into the spatial transformation module to improve the robustness of features.

\begin{figure*}[tb]
\begin{center}
\includegraphics[width=\linewidth]{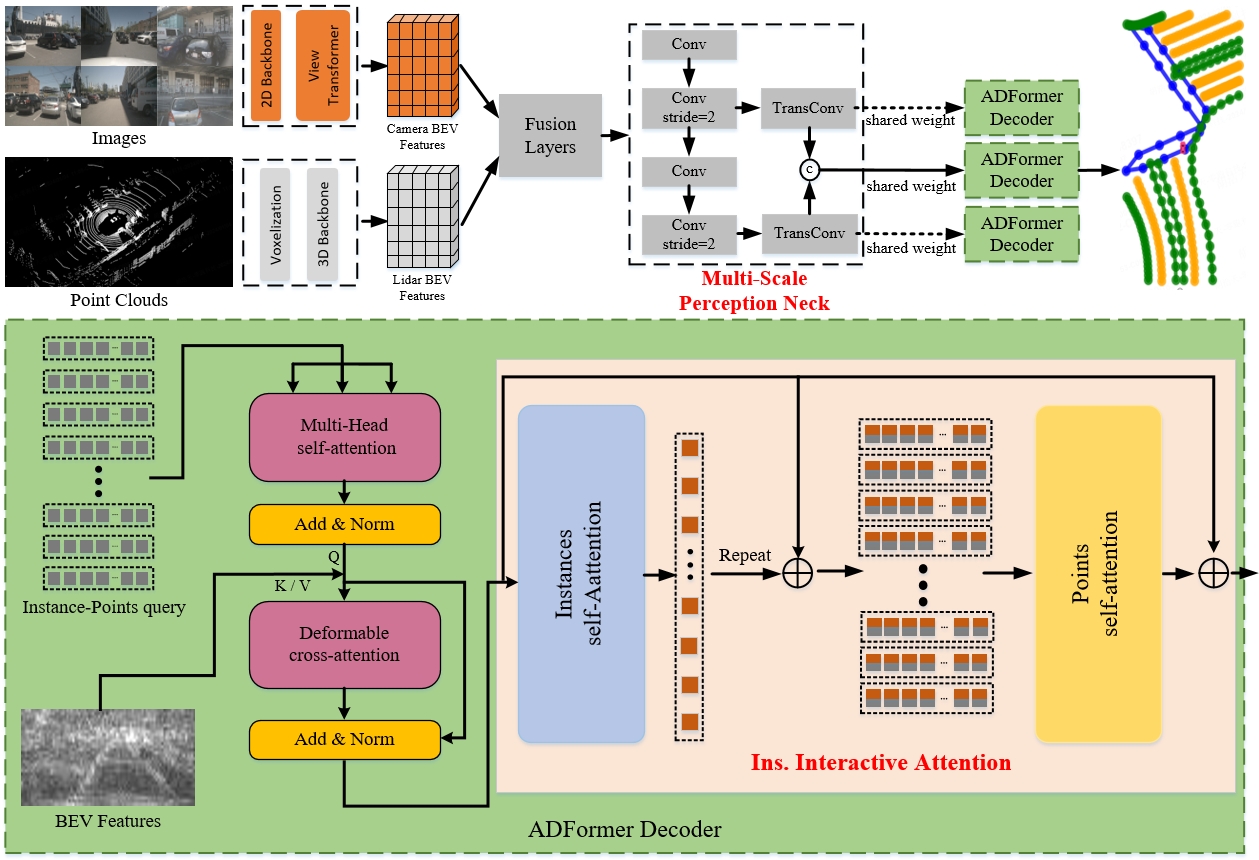}
\end{center}
   \caption{Schematic diagram of the overall framework of ADMap. The figure displays the MPN and IIA proposed in this paper. The process is performed only during training when indicated by the dashed line, and during both training and inference when indicated by the solid line. In decoder, Instance-Points query are defined to represent the topology of the map, and self-attention and cross-attention are used to interact with the BEV map. Instance self-attention and Points self-attention further interact with inter- and intra-instance information.} % ADMap整体框架示意图。图中标红模块为本文所提出的MPN和IIA。在MPN中，虚线部分表示该过程仅在训练中进行，而实线表示该过程在训练和推理中都进行。decoder中定义了若干个hierarchical query来表示地图拓扑结构，self-attention和cross-attention被用于与bev map进行交互，Instance self-attention和Points self-attention进一步交互了实例间和实例内的嵌入。
   \label{fig:Overview}
\end{figure*}
\subsection{HD map construction} \label{map_construction}
% HD map重建
% 传统在线局部地图大多基于多个透视视角的语义分割，使用IPM投影和拼接等方式转换为鸟瞰图视角(BEV)，再经过比较复杂的后处理得到最终需要的地图要素。HDMapNet【】从生成的语义分割结果，实例嵌入和direction prediction中重建了HD Map，然而其需要经过复杂的后处理且精度较低。VectorMapNet【】实现了端到端的HD map重建，其将重建任务与目标检测范式对齐，并通过Polyline Generator逐个生成有序点集。MapTR【】直接使用基于Deformable DETR的目标检测方式生成最终的地图要素，并在模型中引入了permutation-equivalent以解决点序匹配时顺序唯一性约束。
%不同于MapTRv2【】,我们的ADMap利用实例嵌入间的特征建模点级别嵌入间的关系。同时，矢量方向差的引入也更细致的监督了点序位置。
Traditional online local maps\cite{prework1, prework2, prework3, prework4, prework5, prework6, lss} are mostly based on semantic segmentation of multiple perspective views, which are converted to bird's eye view (BEV) using IPM projection and splicing, and then go through a complicated post-processing to get the map elements. HDMapNet \cite{li2021hdmapnet} constructs the HD map from the generated semantic segmentation results, instance embedding, and direction prediction. However, it requires complicated post-processing and has low accuracy. VectorMapNet \cite{liu2022vectormapnet} achieves end-to-end HD map construction by aligning the construction task with the target detection paradigm and generating ordered point sets one by one through Polyline Generator. MapTR \cite{MapTR} generates map elements directly using the Deformable DETR-based object detection paradigm. It introduces permutation-equivalent in the model to address the order uniqueness constraints during point sequence matching.

ADMap utilizes features between instance embeddings to model the relationships between point-level embeddings. Also, the introduction of vectorial direction differences also supervises the point order position more finely.

\section{Method}
\subsection{Overview}
% 概述
% 如图2所示，ADMap的输入为多视角图像和点云。多视角图像$I=\left \{ I_{1},..., I_{M}  \right \} $ 被放入2D backbone中提取图像特征，随后经过LSS-base的View Transformer将2D图像转换到BEV视角，得到camera BEV features $F_{cam}\in  R^{H\times W\times C}$ 。点云$P\in  R^{n\times C}$ 经过体素化后放入3D backbone中，得到lidar BEV features $F_{lidar}\in  R^{H\times W\times C}$ 。Fusion Layers 将camera BEV features和lidar BEV features融合为fusion BEV features $F_{fusion}\in  R^{H\times W\times C}$， 随后放入multi-scale perception neck(3.2节)，其自上到下的融合了BEV features的多尺度特征，以保证网络能准确预测场景中不同尺度的实例。我们在decoder中增加了instance interactive attention(3.3节)，其通过提取的实例嵌入帮助网络更好地捕捉点级别间的关联。此外，vector direction difference loss的细节在3.4节中被介绍。
As shown in Figure~\ref{fig:Overview}, ADMap takes a multiview image and a point cloud as inputs. The features of the multi-view image $I=\left \{ I_{1},..., I_{M}  \right \} $ are extracted by the 2D backbone. The 2D image is then converted to a BEV perspective by LSS-base view transformer so as to obtain camera BEV features $F_{cam}\in  R^{H\times W\times C}$. The point cloud $P\in  R^{n\times C}$ is voxelized and placed into a 3D backbone to obtain lidar BEV features $F_{lidar}\in  R^{H\times W\times C}$. Fusion Layers fuses camera BEV features and lidar BEV features into fusion BEV features $F_{fusion}\in  R^{H\times W\times C}$. Then fusion BEV features inputs into the multi-scale perception neck (Section 3.2), which fuses the multi-scale BEV features from top to bottom. This ensures that the network can predict instances of different scales in the scene accurately. We added instance interactive attention (Section 3.3) in decoder, which helps the network to better capture associations between point levels through extracted instance embeddings. Furthermore, the details of vector direction difference loss are presented in Section 3.4.

\begin{figure*}[tb]
\begin{center}
\includegraphics[width=\linewidth]{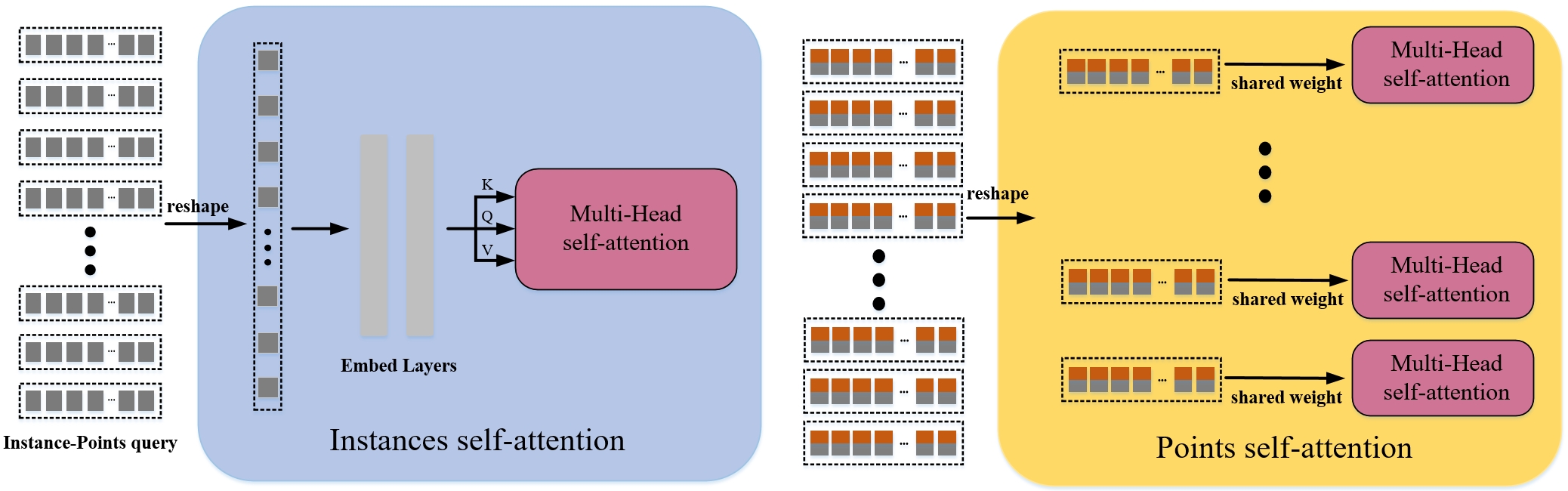}
\end{center}
   \caption{Schematic diagram of Instance self-attention and Points self-attention. The point and channel dimensions in the Instance-Points query are merged and put into embedded layers consisting of multiple MLPs to compress the dimensions. These query are then used in multi-head self-attention for instance-level interactions. Groups the output query of instance self-attention so that each instance is fed into multi-head self-attention separately for point-level interaction.} % Instance self-attention示意图。Instance-Points query的点维度和通道维度合并后被放入由多层MLP组成的embed layers中以维度压缩，随后放入multi-head self-attention中进行实例级别的交互。 随后将Instance self-attention的输出query分组，使每个实例单独输入到multi-head self-attention中以进行点级别的交互。
   \label{fig:Instance_point}
\end{figure*}

\subsection{Multi-scale Perception Neck}\label{subsection}
% multi-scale perception neck
% 传统的FPN结构仅输出融合后的多尺度特征，这使得网络无法细致的捕捉到各级尺度的信息。为了使网络可以获得更为细致的BEV特征，我们提出了multi-scale perception neck(MPN)，其输入为fusion BEV features $F_{fusion}\in  R^{H\times W\times C}$。经过下采样后,各层级BEV特征会各自连接一个上采样层使特征图恢复原来的尺寸，最终各层级特征图合并为多尺度BEV特征$F_{mc}\in  R^{H\times W\times C} $。如图2中的虚线代表该步骤仅在训练时实施，实现代表训练和推理过程中都会实施该步骤。在训练中，多尺度BEV特征和每一层级的BEV特征图都会被送入Transformer decoder进行反向推理，这使得网络可以在不同尺寸预测场景的实力信息以更精细的捕捉多尺度特征。在推理过程中，MPN仅输出多尺度BEV特征$F_{mc}\in  R^{H\times W\times C} $，不会输出各层级特征图，这保证了MPN在推理时占用的资源不变。
Conventional FPN structures only output fused multi-scale features, which prevents the network from delicately capturing information at all levels of scale. To make more detailed BEV features available to the network, we propose multi-scale perception neck (MPN), whose inputs are fusion BEV features  $F_{fusion}\in  R^{H\times W\times C}$. After downsampling, the BEV features of each layer are connected to the upsampled layer to restore its original size. The final feature map of each layer will be merged into a multi-scale BEV feature $F_{mc}\in  R^{H\times W\times C} $. Figure~\ref{fig:Overview} shows that the dotted line indicates that the step is only implemented during training, while the solid line indicates that the step is implemented during both training and inference. During training, the multi-scale BEV features and the BEV feature maps at each level are fed into the ADFormer decoder, which allows the network to predict the instance information of the scene in different scales. During the inference process, MPN only outputs multi-scale BEV features $F_{mc}\in  R^{H\times W\times C} $ and not the feature maps at each level, which ensures that the model inference speed remains constant.

%\begin{figure}[tb]
%\begin{center}
%\includegraphics[width=8.5cm]{Figure3_2.png}
%\end{center}
%   \caption{Schematic of points self-attention. It groups the output query of instance self-attention so that each instance is fed into multi-head self-attention separately for point-level %interaction.}  % points self-attention示意图。其将Instance self-attention的输出query分组，使每个实例单独输入到multi-head self-attention中以进行点级别的交互。
%   \label{fig:Points}
%\end{figure}

\subsection{ADFormer Decoder}
% ADFormer Decoder中定义了一组实例级别的查询$q_{ins}\in R^{N_{i}\times C }$ 和一组点级别的查询$q_{pos}\in R^{N_{p}\times C }$， 随后我们将点级别查询共享到所有实例中，这些分层查询$q$被定义为：
A set of instance-level queries $q_{ins}\in R^{N_{i}\times C }$ and a set of point-level queries $q_{pos}\in R^{N_{p}\times C }$ are defined in the decoder, and we subsequently share the point-level queries across all instances, and Instance-Points queries $q$ are defined as:
\begin{equation}
\label{eq:query}
q=q_{ins}+q_{pos}
\end{equation}

%解码器包含几个级联的解码层，这些层迭代的更新分层查询。在各解码层中，分层查询被输入到self-attention中，这使得分层查询间可以相互交换信息，Deformable Attention被用来对分层查询和多尺度BEV特征$F_{mc}\in  R^{H\times W\times C} $进行交互。
The decoder contains several cascading decoding layers that iteratively update the Instance-Points queries $q$. In each decoding layer, these queries are fed into self-attention, which allows the Instance-Points queries to exchange information with each other. Deformable attention is used to facilitate interaction between the Instance-Points queries and the multiscale BEV features $F_{mc}\in  R^{H\times W\times C} $.

\subsubsection{Instance Interactive Attention}
%为了在解码阶段更好的获取各实例特征，我们提出了instance interactive attention(IIA),其由Instance Self-Attention和Point Self-Attention组成。不同于并行提取实例级和点级别的嵌入【MapTRv2】，我们的IIA级联的提取了点序嵌入，实例嵌入间的特征交互进一步帮助网络学习实例点间的拓扑关系。
%如图3所示，Deformable Cross-Attention输出的分层嵌入$F_{hie} \in   R^{(N_{i} *N_{p})\times C }  $被输入到Instance Self-Attention。将点维度与通道维度合并后，分层嵌入的维度变换为$F_{hie} \in   R^{N_{i}\times (C * N_{p}) }$。随后，分层嵌入被放入多个MLP组成的Embed Layer中获得实例查询，该查询被放入Multi-head self-attention中来捕捉实例间的拓扑关系。得到实例嵌入$F_{ins}\in  R^{N_{i} \times C} $。为了在点级嵌入中融入实例级别信息，我们将实例嵌入$F_{ins}\in  R^{(N_{i} * 1) \times C}$ 和分层嵌入$F_{hie}\in  R^{(N_{i} * N_{p} ) \times C} $相加。相加后的特征被输入值Point Self-Attention中，其对各实例内的店进行交互，进一步精细关联了点序间的拓扑关系。
In order to better acquire the features of each instance in the decoding stage, we propose instance interactive attention (IIA). IIA comprises Instance self-attention and Point self-attention. Unlike the parallel extraction of instance-level and point-level embeddings\cite{maptrv2}, our IIA cascades to extract point sequence embeddings, feature interactions between instance embeddings assist the network in learning the topological relationships between instance points.

Figure~\ref{fig:Instance_point} shows that the hierarchical embedding $F_{hie} \in   R^{(N_{i} *N_{p})\times C }  $ produced by cross-attention is inputted into Instance self-attention. After merging the point dimension with the channel dimension, the dimension transformation of the hierarchical embedding is $F_{hie} \in   R^{N_{i}\times (C * N_{p}) }$. Subsequently, the hierarchical embedding is put into Multilayer Perceptrons (MLPs) to obtain the instance features, which is put into multi-head self-attention to capture the topological relations among instances. The instance embedding $F_{ins}\in  R^{N_{i} \times C} $ is obtained. To include instance-level information in the point-level embedding, we add the instance embedding $F_{ins}\in  R^{(N_{i} * 1) \times C}$ to the hierarchical embedding $F_{hie}\in  R^{(N_{i} * N_{p} ) \times C} $. The summed features are fed into Point self-attention, which interacts with the points within each instance to further finely correlate the topological relationships between point sequences.

\subsubsection{Vector Direction Difference Loss}
%高精地图中包含了矢量化的静态地图元素，包括车道线/路沿和人行横道等。我们针对这些开放形状(车道线/路沿)和封闭形状(人行横道)提出了 vector direction difference loss(VDDL)。通过建模实例内部的点序矢量方向，预测矢量方向和真实矢量方向的差值被用于更细致的监督点序位置。此外，真实矢量方向差较大的点被认为代表了部分场景拓扑的剧烈变化（更不容易预测），需要给予更多关注，以保证网络可以准确预测到这些剧烈变化的点。
%如图4所示，我们首先建模预测点序${\{P_{i,j}^{pre}\}}_{i=0,j=0}^{N_{i} ,N_{p}}$和真实点序${\{P_{i,j}^{gt}\}}_{i=0,j=0}^{N_{i} ,N_{p}}$中的预测矢量线${\{L_{i,j}^{pre}\}}_{i=0,j=0}^{N_{i} ,N_{p}-1 }$和真实矢量线${\{L_{i,j}^{gt}\}}_{i=0,j=0}^{N_{i} ,N_{p}-1}$。
%为了保证相反的角度不会得到相同的损失，我们计算矢量线角度差余弦值cos\theta _{i,j}^{pre}:
HD maps contain vectorized static map elements, including lane lines, road boundaries, and pedestrian crossings. We propose vector direction difference loss (VDDL) for these open shapes (lane lines, road boundaries) and closed shapes (pedestrian crossings). By modeling the point sequence vector direction inside the instance, the difference between the predicted vector direction and the real vector direction is used for finer supervision of the point sequence position. Furthermore, it is believed that points with significant differences in real vector direction indicate significant changes in the topology of certain parts of the scene, which are less predictable. Therefore, more attention is required to ensure that the network can accurately predict these points of drastic change.

\begin{figure}[t]
\begin{center}
\includegraphics[width=8.5cm]{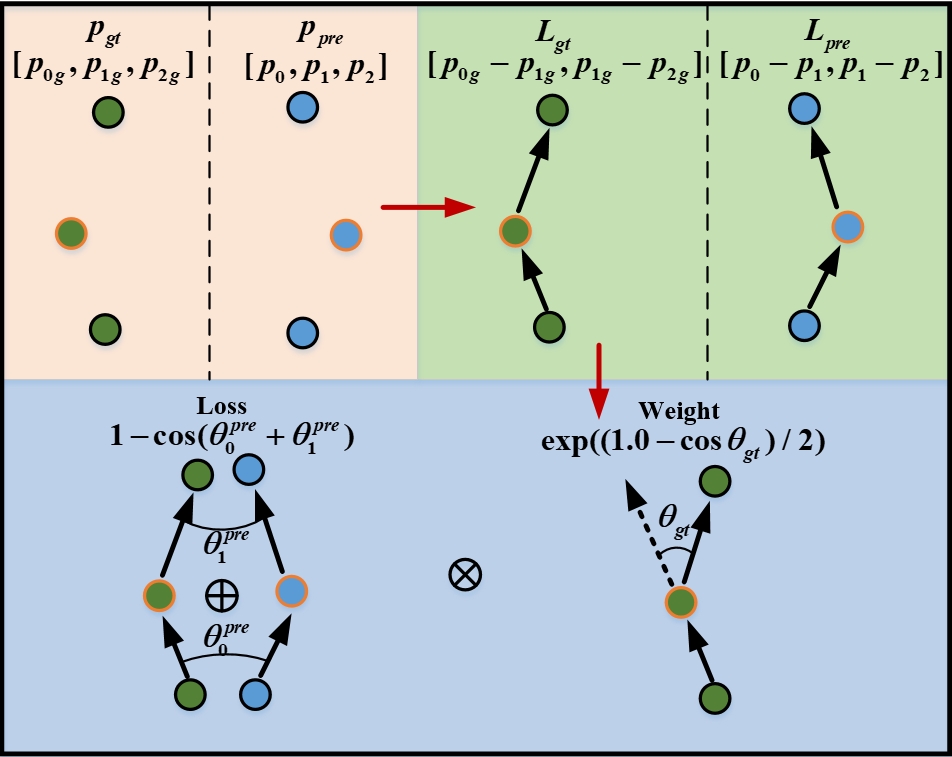}
\end{center}
   \caption{Flowchart of VDDL. The point sequence $P$ is modeled as a vector line $L$ and the vector direction difference between the predicted and ground truth is calculated. The weights of each instance point are obtained from the geometric topological relations of ground truth.} % vector direction difference loss流程图。将点序$P$建模为矢量线$L$，并计算预测值和真实值之间的矢量方向差。通过真值的几何拓扑关系计算各点权重并相乘。
   \label{fig:loss}
\end{figure}

Figure~\ref{fig:loss} illustrates the modeling of the predicted vector line $\{ {L_{i,j}^{pre}\}}_{i=0,j=0}^{N_{i} ,N_{p}-1 }$ and the real vector line ${\{L_{i,j}^{gt}\}}_{i=0,j=0}^{N_{i} ,N_{p}-1}$ in the predicted point sequence ${\{P_{i,j}^{pre}\}}_{i=0,j=0}^{N_{i} ,N_{p}}$ and the real point sequence ${\{P_{i,j}^{gt}\}}_{i=0,j=0}^{N_{i} ,N_{p}}$.

To ensure that the loss increases with the angular difference, we compute the vector line angle difference cosine $cos\theta _{i,j}^{pre}$:

\begin{equation} \label{eq:cos1}
cos\theta _{i,j}^{pre} = 
\left\{
\begin{array}{ll}
 \frac{sum(L_{i,j}^{pre}\ * \ L_{i,j}^{gt})}{norm(L_{i,j}^{pre}) * norm(L_{i,j}^{gt})},\  j\ne N_{p},\\ 
 \ \ \ \ \ \ \ \ \ \ \ \ \ \ \ \ 0, \  \ \ \ \ \ \ \ \ \ \ \ \ \ \ \ \ \ \ j=N_{p},  \\
\end{array}\right.
\end{equation}

The $sum()$ accumulates the 2D coordinate positions of the vector lines, while the $norm()$ normalizes them. We assign weights of different magnitudes to the vector angular differences of the points in the real example. The weights $W_{i,j}^{vec} $ are defined as follows:
%其中sum函数累加了矢量线的二维坐标位置，norm函数归一化了二维坐标位置。我们利用真实实例中各点的矢量角度差来为它们赋予不同大小的权重。权重W_{i,j}^{vec} 定义如下：
\begin{equation} \label{eq:cos2}
cos\theta _{i,j}^{gt} = 
\left\{
\begin{array}{ll}
\frac{sum(L_{i,j-1}^{gt}\ *\ L_{i,j}^{gt})}{norm(L_{i,j-1}^{gt})\ *\ norm(L_{i,j}^{gt})},\ j\ne N_{p} \\ 
 \ \ \ \ \ \ \ \ \ \ \ \ \ \ 0, \  \ \ \ \ \ \ \ \ \ \ \ \ \ \ \ \ j=0 \ or \ N_{p} , \\
\end{array}\right.
\end{equation}

\begin{equation} \label{eq:weight}
W_{i,j}^{vec} = 
\left\{
\begin{array}{ll}
 exp(\frac{1.0-cos\theta _{i,j}^{gt} }{2} ),\ i\ne0\ and\ i\ne N_{p} \\ 
 \ \ \ \ \ \ \ \ \ \ \ \ \ 1.0, \  \ \ \ \ \ \ \ \ \ \ \ i=0 \ or\ i=N_{p}  , \\
\end{array}\right.
\end{equation}

where $N_{p}$ represents the number of points in the instance, and $exp()$ represents an exponential function with base e. The weight of the first and last points is set to 1 since they cannot compute the vector angular difference. When the vector angular difference in the ground truth becomes larger, we assign a larger weight to the point, which makes the network more attentive to significantly varying map topology. The vector direction difference loss $L_{i,j}^{vec}$ is:
%其中N_{p}代表实例中点的数量，$exp()$代表底数为e的指数函数。由于首尾两点无法计算矢量角度差，因此我们将首尾点的权重设置为1。当真实值中的矢量角度差变大时，我们赋予该点更大的权重，这使得网络更为关注显著变化的地图拓扑结构。vector direction difference loss为：
\begin{equation}
\begin{split}
\label{eq:loss}
L_{i,j}^{vec} =\sum_{i=0}^{N_{i} }  \sum_{j=0}^{N_{p}-1}(1 - cos\theta _{i,j}^{pre} )*W_{i,j}^{vec} + \\
\sum_{i=0}^{N_{i} }  \sum_{j=1}^{N_{p}}(1 - cos\theta _{i,j}^{pre} )*W_{i,j}^{vec}
\end{split}
\end{equation}
% 我们使用$1 - cos\theta$将损失的区间调整为$[0,2]$。通过将各点的相邻矢量线角度差余弦相加，该损失更全面的涵盖了各点的几何拓扑信息。由于首尾两点仅有一根相邻矢量线，因此首尾两点的损失为单个矢量角度差的余弦值。 
We use $1 - cos\theta$ to adjust the interval of loss by $[0,2]$. By summing the cosines of the angle differences of neighboring vector lines at each point, the loss captures more comprehensive information about the geometric topology of each point. The loss at the first and last points is the cosine of the angular difference between the unique neighboring vectors.

\section{Experiments}
\begin{table*}[h!] 
\small
\centering 
\caption{Results of ADMap in nuScenes benchmark compared to each state-of-the-art method. We performs validation in both camera-only and multi-modal frameworks with 24 and 110 training epochs, respectively. ADMap uses MapTR as the baseline, while ADMapv2 uses MapTRv2 as the baseline. FPS is measured on NVIDIA RTX 3090 GPU with batch size of 1. 'C' denotes the use of camera, and 'L' denotes the use of lidar. Bolding indicates best performance. $\dagger$  represents the addition of EA-LSS~\cite{hu2023ealss}, CBGS~\cite{zhu2019class}, multi-task training and detection pre-training. } % ADMap在nuScenes基准中与各先进方法对比结果。本文分别在camera-only和多模态框架中进行验证，训练epoch被设置为24和110。ADMap使用MapTR作为baseline，ADMapv2使用MapTRv2作为baseline。C表示使用camera，L表示使用lidar。
\begin{tabular}{cccc|ccc|cc} \hline
\label{tab:nuscenes}
Model & Modality & Backbone & Epoch & $AP_{div}$ & $AP_{ped}$ & $AP_{bou}$ & mAP & FPS   \\ \hline
HDMapNet \cite{li2021hdmapnet} & C & EB0 & 30 & 27.7 & 10.3 & 45.2 & 27.7 & 0.9 \\
HDMapNet \cite{li2021hdmapnet} & C \& L & EB0 \& PP & 30 & 16.3 & 29.6 & 46.7 & 31.0  & 0.5\\
VectorMapNet \cite{liu2022vectormapnet} & C & R50 & 110 & 42.5 & 51.4 & 44.1 & 46.0 & 2.2 \\
VectorMapNet \cite{liu2022vectormapnet} & C \& L & R50 \& PP & 110 & 48.2 & 60.1 & 53.0 & 53.7 & - \\
PivotNet \cite{ding2023pivotnet} & C & R50 & 24 & 56.2 & 56.5 & 60.1 & 57.6 & 6.7\\
StreamMapNet \cite{Wang2024StreamMapNet} & C & R50 & 30 & 56.9 & 55.9 & 61.4 & 58.1 & 14.2\\
BeMapNet \cite{qiao2023bemapnet} & C & R50 & 30 & 62.3 & 57.7 & 59.4 & 59.8 & 4.2 \\ \hline

MapTR \cite{MapTR} & C & R50 & 24 & 51.5 & 46.3 & 53.1 & 50.3 & \textbf{15.1} \\
MapTR \cite{MapTR} & C \& L  & R50 \& SEC & 24 & 55.9 & 62.3 & 69.3 & 62.5 & 6.0\\
ADMap & C & R50 & 24 & 56.2 & 49.4 & 57.9 & 54.5 & 14.8\\
ADMap & C \& L & R50 \& SEC & 24 & 66.6 & 63.3 & 74.0 & 68.0 & 5.8 \\
ADMap & C \& L & R50 \& SEC & 110 & 66.5 & 71.2 & 76.9 & 71.5 & - \\ \hline

MapTRv2 \cite{maptrv2} & C & R50 & 24 & 59.8 & 62.4 & 62.4 & 61.5 & 14.1 \\
ADMapv2 & C & R50 & 24 & 61.9 & 63.5 & 63.3 & 62.9 & 14.8 \\
MapTRv2 \cite{maptrv2} & C \& L & R50 \& SEC & 24 & 65.6 & 66.5 & 74.8 & 69.0 & 5.8 \\
ADMapv2 & C \& L & R50 \& SEC & 24 &68. & 69.0 & 75.2 & 70.8 & 5.8 \\
ADMapv2 & C \& L & R50 \& SEC & 110 & 67.7 & 73.8 & 76.6 & 72.7 & - \\
ADMapv2$\dagger$ & C \& L & R50 \& SEC & 24 & \textbf{83.0} & \textbf{80.2} & \textbf{84.8} & \textbf{82.8} & 5.7 \\
\hline
\end{tabular}
\end{table*}

\begin{table*}[h!] 
\small
\centering 
\caption{Results of ADMap in Argoverse2 benchmark compared to each state-of-the-art method. ADMap uses MapTR as the baseline, while ADMapv2 uses MapTRv2 as the baseline. FPS is measured on NVIDIA RTX3090 GPU with batch size of 1. 'C' denotes the use of camera, and 'L' denotes the use of lidar. * indicates the replicated results in the same framework. } % ADMap在Argoverse2基准中与各先进方法的对比结果。ADMap使用MapTR作为baseline，ADMapv2使用MapTRv2作为baseline。C表示使用camera，L表示使用lidar。*表示在相同框架下的复现结果。i

\begin{tabular}{cccc|ccc|cc} \hline
\label{tab:argoverse2}
Model & Modality & Backbone & Epoch & $AP_{div}$ & $AP_{ped}$ & $AP_{bou}$ & mAP & FPS   \\ \hline
HDMapNet \cite{li2021hdmapnet} & C & EB0 & - & 13.1 & 5.7 & 37.6 & 18.8 & - \\
VectorMapNet \cite{liu2022vectormapnet} & C & R50 & - & 38.3 & 36.1 & 39.2 & 37.9 & - \\ \hline
MapTR* \cite{MapTR} & C & R50 & 6 & 65.5 & 56.6 & 61.8 & 61.3 & \textbf{13.0}\\
ADMap & C & R50 & 6 & 68.9 & 60.3 & 64.9 & 64.7 & 12.6 \\
ADMap & C \& L & R50 \& SEC & 6 & 75.5 & 69.5 & 80.5 & 75.2 & 8.9\\ \hline
MapTRv2 \cite{maptrv2} & C & R50 & 6 & 62.9 & 72.1 & 67.1 & 67.4 & 12.1 \\
ADMapv2 & C & R50 & 6 & 72.4 & 64.5 & 68.9 & 68.7 & 12.6 \\
\textbf{ADMapv2} & \textbf{C \& L} & \textbf{R50 \& SEC} & \textbf{6} & \textbf{76.2} & \textbf{72.8} & \textbf{81.5} & \textbf{76.9} & 8.9 \\ \hline
\end{tabular}
\end{table*}
% 表2报告了ADMap在Argoverse2中的指标
\begin{table*}[h!] 
\centering 
\caption{Ablation experiments for each module. IIA for instance interactive attention, MPN for multi-scale perception neck, and VDDL for vector direction difference loss.}  % 各模块消融实验，IIA表示instance interactive attention，MPN表示multi-scale perception neck，VDDL表示vector direction difference loss。

\begin{tabular}{ccc|ccc|c}
\hline
\label{tab:abl}
IIA & MPN & VDDL & $AP_{div}$ & $AP_{ped}$ & $AP_{bou}$ & mAP  \\
\hline
\usym{2715} & \usym{2715} & \usym{2715}  & 62.3 & 56.2 & 69.1 & 62.5  \\
\usym{2713} & \usym{2715} & \usym{2715} & 63.3 & 59.5 & 73.3 & 65.4 \\
\usym{2715} & \usym{2715} & \usym{2713} & 64.8 & 56.8 & 72.2 & 64.6 \\
\usym{2713} & \usym{2713}& \usym{2715} & 66.0 & 62.9 & 73.6 & 67.3  \\
\usym{2713} & \usym{2715}& \usym{2713} & 65.6 & 59.7 & 74.2 & 66.5  \\
\usym{2713} & \usym{2713} & \usym{2713} & 66.6 & 63.3 & 74.0 & \textbf{68.0}  \\
\multicolumn{3}{c|}{Improvment} & +4.3 & +7.1 & +4.9 & +5.5\\
\hline
\end{tabular}
\end{table*}

\begin{figure*}[t]
\begin{center}
\includegraphics[width=\linewidth]{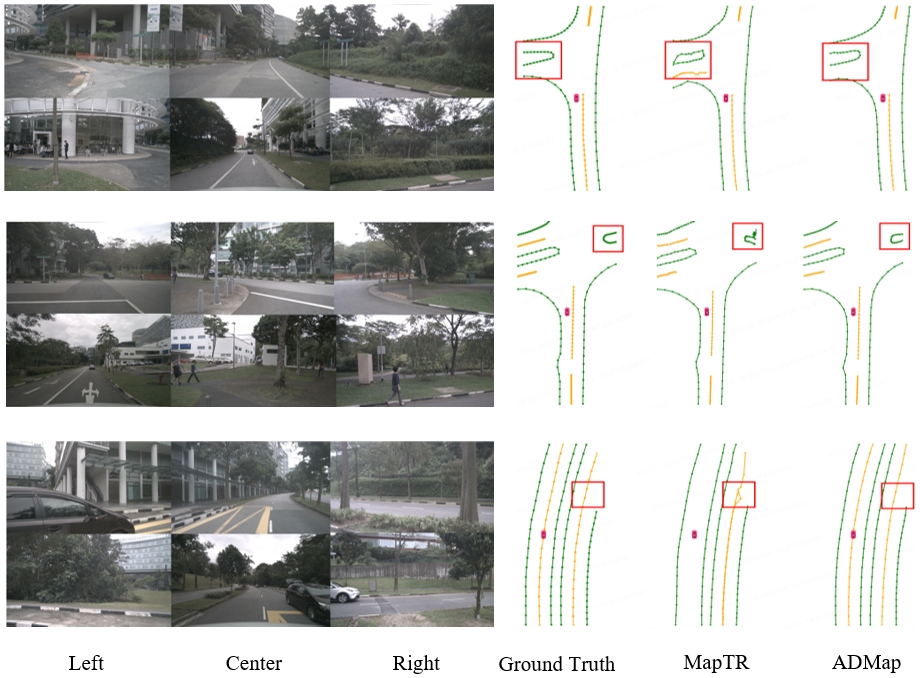}
\end{center}
   \caption{Visualization results of the nuScenes dataset. Areas of discrepancy are indicated by red boxes. ADMap effectively reduces jitter within instances.} % nuScenes 可视化结果
   \label{fig:vis_nus}
\end{figure*}

\subsection{Dataset and metric}
% 数据集和指标
% 我们在nuScenes和Argoverse2中验证了ADMap的有效性，nuScenes数据集中包含了40k个标注数据，关键帧的采样频率为2HZ，其包含了FOV为360度的6个环视摄像头和1个激光雷达。Argoverse2包含了1000个带标注的多模态数据序列，其包含了7个环视摄像头和2个立体摄像机的高分辨率图像，还包含了激光雷达点云/姿态对齐的地图等。
%为了公平的评估ADMap，我们将地图元素分为车道线/道路边界和人行横道三类，采用平均精度(AP)来评估地图构建的质量，使用预测点序和真实点序的chamfer距离之和判断两者是否匹配。chamfer距离阈值设置[0.2, 0.5, 1.0]和[0.5, 1.0, 1.5]，我们分别在这三种阈值下计算AP，并将平均值作为最终指标。
The effectiveness of ADMap was verified in nuScenes \cite{nuscenes2019} and Argoverse2 \cite{Argoverse2} datasets. The nuScenes dataset includes 40,000 labeled data with keyframes sampled at 2 HZ. It consists of 6 surround view cameras and 1 lidar with a FOV of 360 degrees. Argoverse2 comprises 1000 annotated multi-modal data sequences. The dataset includes high-resolution images from 7 surround view cameras and 2 stereo cameras, as well as LIDAR point cloud and attitude-aligned maps.

To ensure a fair evaluation of ADMap, we classify the map elements into three categories: lane lines, road boundaries and pedestrian crossings. We use Average Precision (AP) to evaluate the quality of the map construction, and using the sum of the chamfer distances of the predicted and real point sequences to determine whether they match or not. Chamfer distance thresholds are set to $[0.5, 1.0, 1.5]$, and we separately under these three thresholds calculate the AP and take the average as the metric.

\subsection{Detail}
% 细节
% ADMap使用8*NVIDIA Geforce RTX A100 GPUs进行训练。公平起见，我们使用单卡3090测试模型的FPS。实验中batch size设置为8，学习率设置为6e-4,使用AdamW优化器以及cosine annealing schedule。点云范围设置为[-15.0, -30.0, -5.0, 15.0, 30.0, 3.0]，voxel size设置为[0.15, 0.15, 0.2]。我们使用ResNet50【】作为camera backbone，使用SECOND【】作为lidar backbone。MPN的输入通道数为256， 输出通道数为[512, 512, 512]。损失函数中，L1 Loss和VDDL的权重分别设置为5.0和1.0。 ADMapv2增加了MapTRv2中的Auxiliary Dense Prediction Loss和Auxiliary One-to-Many Set Prediction Loss。为了加快训练速度，Auxiliary One-to-Many Set Prediction Loss中的复制倍率设置为3(Maptrv2中为6)，这会使得模型性能有所下降。
ADMap is trained using 8* NVIDIA Geforce RTX A100 GPUs. The batch size is set to 8, the learning rate is set to 6e-4, and the AdamW optimizer as well as the cosine annealing schedule are used. The point cloud range is set to $[-15.0, -30.0, -5.0, 15.0, 30.0, 3.0]$, and the voxel size is set to $[0.15, 0.15, 0.2]$. We use ResNet50 \cite{resnet2015} as the camera backbone and SECOND \cite{second} as the lidar backbone. The number of input channels of MPN is 256 and the number of output channels is [512, 512, 512]. In the loss function, the weights of L1 Loss and VDDL are set to 5.0 and 1.0, respectively. ADMapv2 increases the Auxiliary Dense Prediction Loss and Auxiliary One-to-Many Set Prediction Loss from MapTRv2~\cite{maptrv2}. To speed up training, the replication multiplier in the Auxiliary One-to-Many Set Prediction Loss is set to 3 (6 in MapTRv2), which decreases the model performance.

\subsection{Comparative experiment}
\subsubsection{nuScenes}

% 表1报告了ADMap和最先进方法在nuScenes数据集的指标。可以看到，不论是camera-only还是多模态框架中，ADMap的效果都要远好于基线。在camera-only框架中，ADMap相较于MapTR提高了4.2\%，相较于MapTRv2提高了()\%。在多模态框架中，ADMap相较于MapTR提高了5.5\%，ADMapv2相较于MapTRv2提高了1.3%，其最高精度为()\%，达到了该基准中的最佳性能。在速度方面,我们的ADMap和ADMapv2也展现出了优秀的性能。相较于baseline，ADMap在大幅提高模型性能的同时，FPS仅下降了0.3和0.2。值得一提的是，ADMapv2相较于baseline不仅提高了性能，推理速度也有所提高。camera-only框架的推理延迟从70.9ms降低为68.9ms。在多模态框架中，由于模型自身延迟较大，因此ADMapv2的速度优势无法体现。

Table~\ref{tab:nuscenes} reports the metrics of ADMap and the state-of-the-art methods on the nuScenes dataset. It is evident that ADMap outperforms the baseline in both camera-only and multimodal frameworks. In the camera-only framework, ADMap improves 4.2\% compared to MapTR, ADMapv2 improves 1.4\% compared to MapTRv2. In the multimodal framework, ADMap improves by 5.5\% compared to MapTR and ADMapv2 improves by 1.8\% compared to MapTRv2, with a maximum accuracy of 72.7\%, which achieves the best performance in this benchmark.

In terms of speed, both ADMap and ADMapv2 demonstrate excellent performance. ADMap significantly improves model performance compared to the baseline, with only a slight decrease in FPS of 0.3 and 0.2. It is worth noting that ADMapv2 not only improves performance but also enhances inference speed compared to the baseline. The inference latency of the camera-only framework is reduced from 70.9ms to 67.6ms. In the multimodal framework, the speed advantage of ADMapv2 cannot be realized due to the model itself has a large latency.

\begin{figure*}[t]
\begin{center}
\includegraphics[width=\linewidth]{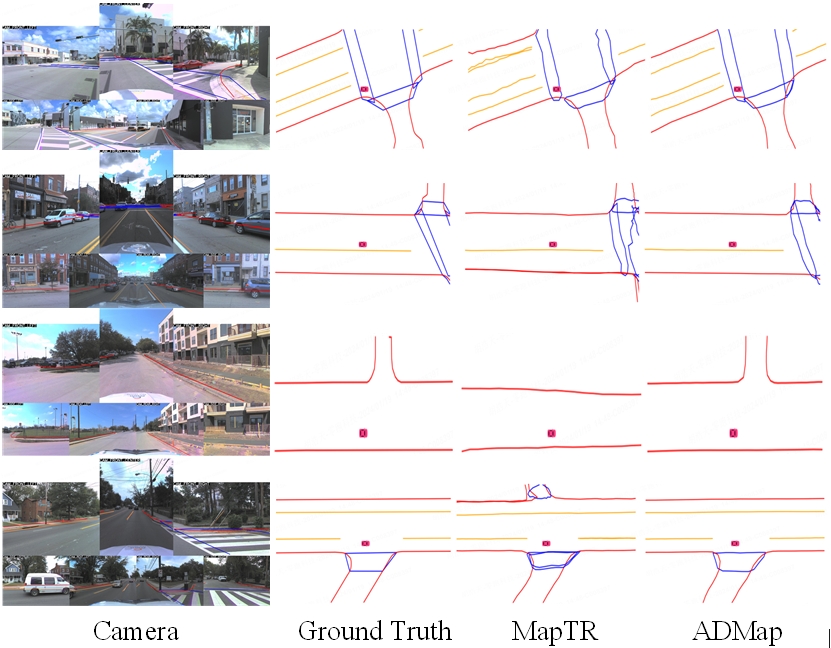}
\end{center}
   \caption{Visualization results of the Argoverse2 dataset. Areas of discrepancy are indicated by red boxes. ADMap effectively reduces jitter within instances.} % Argoverse2 可视化结果
   \label{fig:vis_av2}
\end{figure*}

\begin{figure*}[t]
\begin{center}
\includegraphics[width=\linewidth]{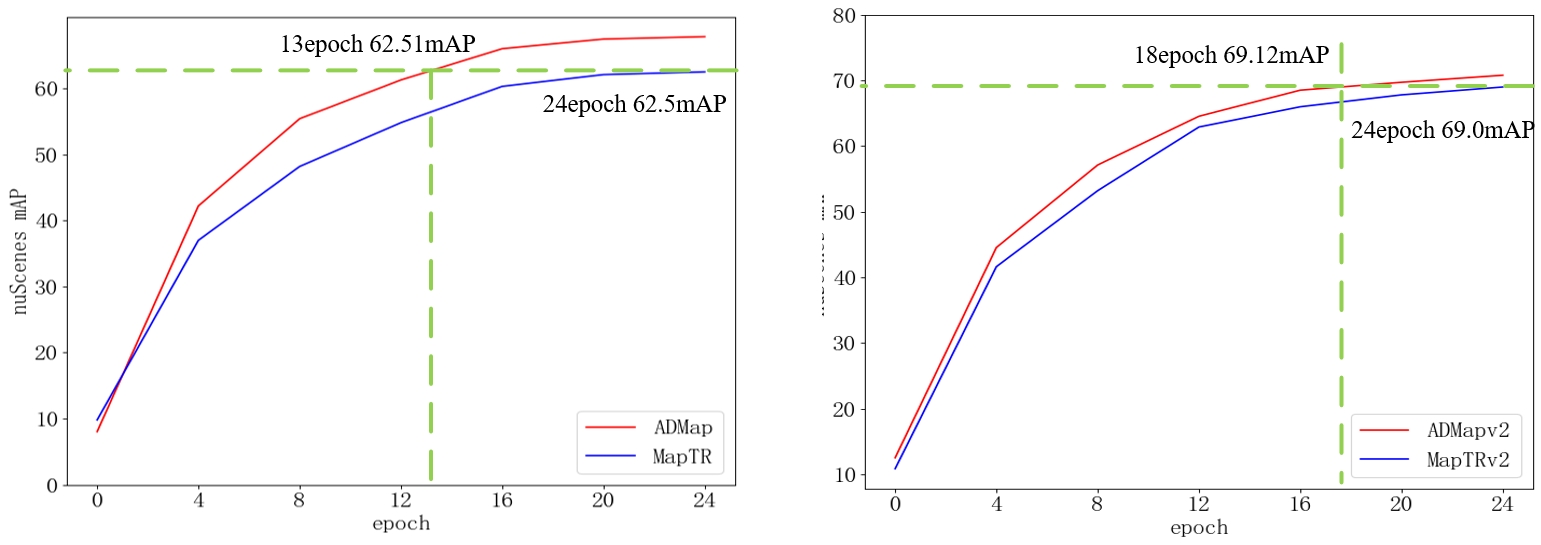}
\end{center}
   \caption{ADMap, MapTR, ADMapv2, and MapTRv2 convergence curves in the nuScenes dataset. ADMap and ADMapv2 exceed their baseline best performance at epoch 13 and epoch 18, respectively} % nuScenes数据集中ADMap、MapTR、ADMapv2和MapTRv2收敛曲线。
   \label{fig:curv}
\end{figure*}

%\begin{table}[h!] 
%\small
%\centering 
%\caption{Ablation experiments for each module. IIA for instance interactive attention, MPN for multi-scale perception neck, and VDDL for vector direction difference loss.}  % 各模块消融实验，IIA表示instance interactive attention，MPN表示multi-scale perception neck，VDDL表示vector direction difference loss。
%\begin{tabular}{c|ccc|c} \hline
%\label{tab:abl}
%Model & $AP_{div}$ & $AP_{ped}$ & $AP_{bou}$ & mAP    \\ \hline
%MapTR baseline & 62.3 & 56.2 & 69.1 & 62.5\\
%+IIA & 63.3 & 59.5 & 73.3 & 65.4 \\
%+MPN & 66.0 & 62.9 & 73.6 & 67.3 \\
%+VDDL & \textbf{66.6} & \textbf{63.3} & \textbf{74.0} & \textbf{68.0} \\ \hline
%\end{tabular}
%\end{table}

\begin{table}[h!] 
\small
\centering 
\caption{Impact of different self-attention on performance. DSA denotes Decoupled self-attention, IIA denotes instance interactive attention.  FPS is measured on NVIDIA RTX3090 GPU with batch size of 1.}  % 不同self-attention对模型性能的影响。DSA表示decoupled self attention, IIA表示instance interactive attention。
\begin{tabular}{c|ccc|cc} \hline
\label{tab:abl_IIA}
Model & $AP_{div}$ & $AP_{ped}$ & $AP_{bou}$ & mAP & FPS   \\ \hline
Vanilla & 62.3 & 56.2 & 69.1 & 62.5 & \textbf{6.0}\\
+DSA \cite{maptrv2} & 64.2 & 57.8 & 70.2 & 64.1 & 5.8\\
+IIA(Our) & 63.3 & 59.5 & 73.3 & \textbf{65.4} & 5.9 \\\hline
\end{tabular}
\end{table}

\begin{table}[h!] 
\small
\centering 
\caption{Impact of inserting the neck structure on the model performance. SECONDNeck denotes the neck structure in SECOND, MPN denotes multi-scale perception neck.} % 插入neck结构对模型性能的影响。BackboneNeck表示SECOND中的neck结构, MPN表示multi-scale perception neckmulti-scale perception neck。
\begin{tabular}{c|ccc|cc} \hline
\label{tab:abt_MPN}
Model & $AP_{div}$ & $AP_{ped}$ & $AP_{bou}$ & mAP &FPS   \\ \hline
Vanilla & 62.3 & 56.2 & 69.1 & 62.5 & \textbf{6.0}  \\
+SECONDNeck \cite{second} & 62.8 & 58.4 & 70.2 & 63.7 & 5.9 \\
+MPN(Our) & 65.0 & 59.6 & 69.4 & \textbf{64.5} & 5.9 \\\hline
\end{tabular}
\end{table}

\begin{table}[h!] 
\centering 
\caption{Impact of VDDL weights on performance} % 
\begin{tabular}{c|ccc|c} \hline
\label{tab:weight}
Weight & $AP_{div}$ & $AP_{ped}$ & $AP_{bou}$ & mAP     \\ \hline
0.0 & 51.5 & 46.3 & 53.1 & 50.3 \\
0.5 & 54.6 & 49.3 & 53.8 & 52.5 \\
\textbf{1.0} &  55.9 & 48.4 & 55.6 & \textbf{53.3}  \\
2.0 &  55.5 & 45.4 & 54.2 & 51.7 \\
3.0 & 55.4 & 45.2 & 55.1 & 51.9 \\\hline
\end{tabular}
\end{table}

\begin{table}[h!] 
\centering 
\caption{Ablation experiments on the number of downsampled layers in MPN.} %
\begin{tabular}{c|ccc|cc} \hline
\label{tab:num_layer}
num layer & $AP_{div}$ & $AP_{ped}$ & $AP_{bou}$ & mAP &FPS   \\ \hline
1 &  66.5 & 68.6 & 74.2 & 69.8 & \textbf{6.0} \\
\textbf{2} & 67.9 & 68.5 & 74.5 & 70.3 & 5.8\\
3 & 68.1 & 68.3 & 74.9 & \textbf{70.5} & 5.3 \\\hline
\end{tabular}
\end{table}

\begin{table}[h!] 
\centering 
\caption{Comparison of average chamfer distance(ACD) between ADMap and ADMapv2 and their respective baselines.}
\begin{tabular}{c|cccc} \hline
\label{fig:chamfer}
model & MapTR & \textbf{ADMap} & MapTRv2 & \textbf{ADMapv2}  \\ \hline
ACD & 0.597 & \textbf{0.518} & 0.560 & \textbf{0.422} \\ \hline
\end{tabular}
\end{table}

\subsubsection{Argoverse2}
%表2报告了在Argoverse2数据集中ADMap与最先进方法的指标。在camera-only框架中，ADMap和ADMapv2相较于baseline分别提高了3.4%和1.3%。在多模态框架中，ADMap和ADMapv2的指标为了75.2%和76.9%，达到了该基准中的最佳性能。在推理速度方面，ADMapv2的延迟相较于MapTRv2下降了11.4ms。
Table~\ref{tab:argoverse2} reports the metrics of ADMap and the state-of-the-art methods on the Argoverse2 dataset. In the camera-only framework, ADMap and ADMapv2 improved by 3.4\% and 1.3\% compared to the baseline, respectively. In the multimodal framework, ADMap and ADMapv2 achieved the best performance in the benchmark with 75.2\% and 76.9\%. Regarding inference speed, the latency of ADMapv2 decreased by 3.3ms compared to MapTRv2.

\subsection{Ablation experiment}
% 各模块在nuscenes基准中的消融实验结果如表3所示。在MapTR的基础上，我们增加了MPN/IIA和VDDL。可以看出，单独增加IIA后mAP提高了2.9\%，单独增加VDDL后，mAP提高了2.1\%。同时增加IIA和MPN后mAP提高了4.8%，同时增加IIA和VDDL后mAP提高了4.0%。最终我们所提出了方法极大提高了基线的能力，使其mAP提高了5.5%。 
% 表4报告了IIA与Decoupled Self-Attention(DSA)对baseline的提升效果对比。可以看出，IIA取得了更好的效果。DSA使基线的mAP提高了2.6%，而IIA使基线的mAP提高了2.9%。此外，IIA在速度方面也优于DSA，其延迟相较于DSA减少了2.9ms。
% 表5报告了增加neck层对mAP的影响。增加了SECOND【】的neck层后，mAP提高了1.2\%，而增加了MPN后，在不增加推理时间的情况下，模型mAP提高了2.0\%

Table~\ref{tab:abl} shows the results of the ablation experiments for each module in the nuscenes benchmark. To MapTR, we added MPN/IIA and VDDL. It can be observed that the mAP increased by 2.9\% with the addition of IIA alone and 2.1\% with the addition of VDDL alone. The mAP increased by 4.8\% after adding both IIA and MPN, and by 4.0\% after adding both IIA and VDDL. Finally, our proposed method greatly improves the ability of the baseline to improve its mAP by 5.5\%.

Table~\ref{tab:abl_IIA} shows the comparison between IIA and Decoupled Self-Attention (DSA) and their respective effects. It can be seen that IIA achieves better results. DSA improves the mAP of the baseline by 1.6\%, while IIA improves the mAP of the baseline by 2.9\%. In addition, IIA also outperforms DSA in terms of speed, and its latency is reduced by 2.9ms compared to DSA.

Table~\ref{tab:abt_MPN} shows how the performance is affected by adding neck. The mAP increased by 1.2\% with the addition of the neck of SECOND\cite{second}, and the model mAP increased by 2.0\% with the addition of MPN, without increasing the inference time.

% 表6报告了VDDL的权重在nuScenes基准中对于camera-only框架的ADMap性能的影响。可以看到，当权重设置为1.0时，mAP达到了53.3%，实现了最佳性能。

%表7报告了MPN的下采样层数在nuScenes基准中对于多模态框架的ADMapv2性能的影响。由于下采样层数的增加会导致模型推理速度变慢，为了兼顾速度与性能，我们将下采样层数设置为2。
Table~\ref{tab:weight} reports the impact of VDDL's weights in the nuScenes benchmark on the performance of ADMap for the camera-only framework. The best performance is achieved when the weights are set to 1.0. 

Table~\ref{tab:num_layer} reports the effect of the number of downsampling layers of the MPN on the performance of ADMapv2 in the nuScenes benchmarks. To balance speed and performance, we set the number of downsampling layers to 2, as increasing the number of downsampling layers can slow down model inference.

\subsection{Visualization}
%图5和图6展示了gt，MapTR和ADMap在nuScenes和Argoverse2中的可视化对比结果。从图中可以看到，MapTR中车道线、人行横道和路边沿等类别实例皆出现了一定程度的歪曲和变形，这会对后续规划、控制等任务造成验证影响。而ADMap有效的缓解了矢量实例内部点的抖动，更准确的预测了地图元素。
Figure~\ref{fig:vis_nus} and Figure~\ref{fig:vis_av2} shows a comparison of the visualizations between ground truth, MapTR, and ADMap in the nuScenes and Argoverse2 benchmark. The figure shows that the lane lines, pedestrian crossings, and road boundaries in MapTR are distorted and deformed to some extent. This distortion can affect subsequent planning, control, and other tasks. In contrast, ADMap effectively mitigates point jitter within vector instances and predicts map elements more accurately.

\subsection{Analystic}

% 在nuScenes基准中，我们挑选了所有内部chamfer distance小于1.5的预测实例，并计算这些预测实例(true positive instance)和真实实例间的平均chamfer distance，以此作为评价指标来评估模型对实例内部点位置的预测准确情况。
%结果如表8所示。当平均chamfer distance越小时，代表预测实例越准确(实例点抖动现象越少)。ADMap的chamfer distance loss相较于MapTR下降了0.079，ADMapv2相较于baseline下降了0.138，这表明ADMap有效缓解了点序预测不准的问题。

For the nuScenes benchmark, we select all predicted instances whose sum of the internal chamfer distance is less than 1.5. We then calculate the average chamfer distance between these predicted instances (true positive instances) and the real instances. This serves as an evaluation metric to assess the accuracy of the point locations inside the instances. We then calculate the average chamfer distance between these predicted instances (true positive instances) and the ground truth. Table~\ref{fig:chamfer} displays the results. A smaller average chamfer distance indicates more accurate instance prediction (less instance point jitter phenomenon). The chamfer distance loss of ADMap decreases by 0.079 compared to MapTR, and that of ADMapv2 decreases by 0.138 compared to baseline. This indicates that ADMap effectively mitigates the problem of inaccurate point sequence prediction.
% 如图7所示，我们在nuScenes基准中对比了ADMap, MapTR, ADMapv2, and MapTRv2的收敛曲线。当ADMap训练到第13个epoch时，其mAP已经超过了MapTR的最优性能，而ADMapv2在训练到第18个epoch时，其mAP到达了69.12，超过了MapTRv2的最佳性能。

Figure~\ref{fig:curv} shows a comparison of the convergence curves of ADMap, MapTR, ADMapv2, and MapTRv2 in the nuScenes benchmark. By the 13th epoch, mAP of ADMap exceeds the optimal performance of MapTR, while ADMapv2 reaches 69.12 mAP by the 18th epoch, exceeding the optimal performance of MapTRv2.

\section{Conclusions}
% ADMap是一个有效且高效的矢量化高清地图重建框架，其通过multi-scale perception neck，instance interactive attention和vector direction difference loss有效缓解了实例点抖动造成的地图拓扑失真问题。大量实验表明了我们提出的方法可以在nuScenes和Argoverse2基准中实现最佳的性能，且其高效的特点也已被验证。我们希望ADMap可以帮助社区推进矢量化高清地图重建任务的研究，以帮助自动驾驶等领域更好的发展。
ADMap is an effective and efficient vectorized HD map construction framework that effectively mitigates the problem of map topology distortion caused by instance point jitter through three modules: multi-scale perception neck, instance interactive attention, and vector direction difference loss. Numerous experiments have shown that our proposed method can achieve the best performance in nuScenes and Argoverse2 benchmarks, and its high efficiency has also been verified. We believe that ADMap can assist the community in advancing research on vectorized HD map construction tasks for better development in areas such as autonomous driving.

{\small
\bibliographystyle{iccv}
\bibliography{iccv}

\begin{thebibliography}{10}\itemsep=-1pt

\bibitem{nuscenes2019}
Holger Caesar, Varun Bankiti, Alex~H. Lang, Sourabh Vora, Venice~Erin Liong, Qiang Xu, Anush Krishnan, Yu Pan, Giancarlo Baldan, and Oscar Beijbom.
\newblock nuscenes: A multimodal dataset for autonomous driving.
\newblock {\em arXiv preprint arXiv:1903.11027}, 2019.

\bibitem{chen2022persformer}
Li Chen, Chonghao Sima, Yang Li, Zehan Zheng, Jiajie Xu, Xiangwei Geng, Hongyang Li, Conghui He, Jianping Shi, Yu Qiao, and Junchi Yan.
\newblock Persformer: 3d lane detection via perspective transformer and the openlane benchmark.
\newblock In {\em European Conference on Computer Vision (ECCV)}, 2022.

\bibitem{prework1}
S. Chen, T. Cheng, X. Wang, W. Meng, Q. Zhang, and W. Liu.
\newblock Efficient and robust 2d-to-bev representation learning via geometryguided kernel transformer.
\newblock {\em arXiv preprint arXiv:2206.04584}, 2022.

\bibitem{ding2023pivotnet}
Wenjie Ding, Limeng Qiao, Xi Qiu, and Chi Zhang.
\newblock Pivotnet: Vectorized pivot learning for end-to-end hd map construction.
\newblock In {\em Proceedings of the IEEE/CVF International Conference on Computer Vision}, pages 3672--3682, 2023.

\bibitem{Gao2024Complementing}
Wenjie Gao, Jiawei Fu, Yanqing Shen, Haodong Jing, Shitao Chen, and Zheng Nanning.
\newblock Complementing onboard sensors with satellite maps: A new perspective for hd map construction.
\newblock {\em ICRA}, 2024.

\bibitem{3dlanenet}
Noa Garnett, Rafi Cohen, Tomer Pe'er, Roee Lahav, and Dan Levi.
\newblock 3d-lanenet: End-to-end 3d multiple lane detection.
\newblock In {\em Proceedings of the IEEE/CVF International Conference on Computer Vision (ICCV)}, October 2019.

\bibitem{guo2020genlanenet}
Yuliang~Guo Guo, Guang Chen, Zhao Peitao, Zhang Weide, Jinghao Miao, Jingao Wang, and Tae~Eun Choe.
\newblock Gen-lanenet: A generalized and scalable approach for 3d lane detection.
\newblock 2020.

\bibitem{resnet2015}
Kaiming He, Xiangyu Zhang, Shaoqing Ren, and Jian Sun.
\newblock Deep residual learning for image recognition.
\newblock {\em arXiv preprint arXiv:1512.03385}, 2015.

\bibitem{prework3}
A. Hu, Z. Murez, N. Mohan, S. Dudas, J. Hawke, V. Badrinarayanan, R. Cipolla, and Kendall A.
\newblock Fiery: Future instance segmentation in bird’s-eye view from surround monocular cameras.
\newblock {\em ICCV}, 2021.

\bibitem{hu2023ealss}
Haotian Hu, Fanyi Wang, Jingwen Su, Yaonong Wang, Laifeng Hu, Weiye Fang, Jingwei Xu, and Zhiwang Zhang.
\newblock Ea-lss: Edge-aware lift-splat-shot framework for 3d bev object detection.
\newblock {\em arXiv preprint arXiv:2303.17895}, 2023.

\bibitem{li2021hdmapnet}
Qi Li, Yue Wang, Yilun Wang, and Hang Zhao.
\newblock Hdmapnet: An online hd map construction and evaluation framework, 2021.

\bibitem{prework4}
Z. Li, W. Wang, H. Li, E. Xie, C. Sima, T. Lu, Y. Qiao, and J. Dai.
\newblock Bevformer: Learning bird’s-eye-view representation from multicamera images via spatiotemporal transformers.
\newblock {\em ECCV}, 2022.

\bibitem{prework6}
Z. Li, W. Wang, H. Li, E. Xie, C. Sima, T. Lu, Y. Qiao, and J. Dai.
\newblock Bevformer: Learning bird’s-eye-view representation from multicamera images via spatiotemporal transformers.
\newblock {\em ECCV}, 2022.

\bibitem{MapTR}
Bencheng Liao, Shaoyu Chen, Xinggang Wang, Tianheng Cheng, Qian Zhang, Wenyu Liu, and Chang Huang.
\newblock Maptr: Structured modeling and learning for online vectorized hd map construction.
\newblock In {\em International Conference on Learning Representations}, 2023.

\bibitem{maptrv2}
Bencheng Liao, Shaoyu Chen, Yunchi Zhang, Bo Jiang, Qian Zhang, Wenyu Liu, Chang Huang, and Xinggang Wang.
\newblock Maptrv2: An end-to-end framework for online vectorized hd map construction.
\newblock {\em arXiv preprint arXiv:2308.05736}, 2023.

\bibitem{liu2022vectormapnet}
Yicheng Liu, Yuan Yuantian, Yue Wang, Yilun Wang, and Hang Zhao.
\newblock Vectormapnet: End-to-end vectorized hd map learning.
\newblock In {\em International conference on machine learning}. PMLR, 2023.

\bibitem{prework5}
Z. Liu, S. Chen, X. Guo, X. Wang, T. Cheng, H. Zhu, Q. Zhang, W. Liu, and Y. Zhang.
\newblock Vision-based uneven bev representation learning with polar rasterization and surface estimation.
\newblock {\em arXiv preprint arXiv:2207.01878}, 2022.

\bibitem{neven2018lanenet}
Davy Neven, Bert~De Brabandere, Stamatios Georgoulis, Marc Proesmans, and Luc~Van Gool.
\newblock Towards end-to-end lane detection: an instance segmentation approach, 2018.

\bibitem{pre5}
Xingang Pan, Jianping Shi, Ping Luo, Xiaogang Wang, and Xiaoou Tang.
\newblock Spatial as deep: Spatial cnn for traffic scene understanding.
\newblock {\em In Proceedings of the AAAI Conference on Artificial Intelligence}, 2018.

\bibitem{lss}
J. Philion and S. Fidler.
\newblock Lift, splat, shoot: Encoding images from arbitrary camera rigs by implicitly unprojecting to 3d.
\newblock {\em ECCV}, 2020.

\bibitem{qiao2023bemapnet}
Limeng Qiao, Wenjie Ding, Xi Qiu, and Chi Zhang.
\newblock End-to-end vectorized hd-map construction with piecewise bezier curve.
\newblock In {\em Proceedings of the IEEE/CVF Conference on Computer Vision and Pattern Recognition (CVPR)}, pages 13218--13228, June 2023.

\bibitem{Wang2024GeMap}
Shuo Wang, Fan Jia, Yucheng Liu, Yingfei an~Zhao, Zehui Chen, Tiancai Wang, Chi Zhang, Xiangyu Zhang, and Feng Zhao.
\newblock Stream query denoising for vectorized hd map construction.
\newblock {\em arXiv preprint arXiv:2401.09112}, 2024.

\bibitem{Argoverse2}
Benjamin Wilson, William Qi, Tanmay Agarwal, John Lambert, Jagjeet Singh, Siddhesh Khandelwal, Bowen Pan, Ratnesh Kumar, Andrew Hartnett, Jhony~Kaesemodel Pontes, Deva Ramanan, Peter Carr, and James Hays.
\newblock Argoverse 2: Next generation datasets for self-driving perception and forecasting.
\newblock In {\em Proceedings of the Neural Information Processing Systems Track on Datasets and Benchmarks (NeurIPS Datasets and Benchmarks 2021)}, 2021.

\bibitem{xie2023mvmap}
Ziyang Xie, Ziqi Pang, and Yu-Xiong Wang.
\newblock Mv-map: Offboard hd-map generation with multi-view consistency.
\newblock {\em arXiv}, 2023.

\bibitem{second}
Yan Yan, Yuxing Mao, and Bo Li.
\newblock Second: Sparsely embedded convolutional detection.
\newblock 2018.

\bibitem{yu2023ScalableMap}
Jingyi Yu, Zizhao Zhang, Shengfu Xia, and Jizhang Sang.
\newblock Scalablemap: Scalable map learning for online long-range vectorized hd map construction.
\newblock {\em CoRL}, 2023.

\bibitem{Wang2024StreamMapNet}
Tianyuan Yuan, Yicheng Liu, Yue Wang, Yilun Wang, and Hang Zhao.
\newblock Streammapnet: Streaming mapping network for vectorized online hd map construction.
\newblock {\em arXiv preprint arXiv:2308.12570}, 2023.

\bibitem{pre6}
Tu Zheng, Hao Fang, Yi Zhang, Wenjian Tang, Zheng Yang, Haifeng Liu, and Deng Cai.
\newblock Resa: Recurrent feature-shift aggregator for lane detection.
\newblock {\em In Proceedings of the AAAI Conference on Artificial Intelligence}, 2021.

\bibitem{prework2}
B. Zhou and P. Krahenbuhl.
\newblock Cross-view transformers for real-time map-view semantic segmentation.
\newblock {\em CVPR}, 2022.

\bibitem{zhu2019class}
Benjin Zhu, Zhengkai Jiang, Xiangxin Zhou, Zeming Li, and Gang Yu.
\newblock Class-balanced grouping and sampling for point cloud 3d object detection.
\newblock {\em arXiv preprint arXiv:1908.09492}, 2019.

\end{thebibliography}
}

\end{document}

\begin{figure*}[t]
\small
\begin{center}
\includegraphics[width=\linewidth]{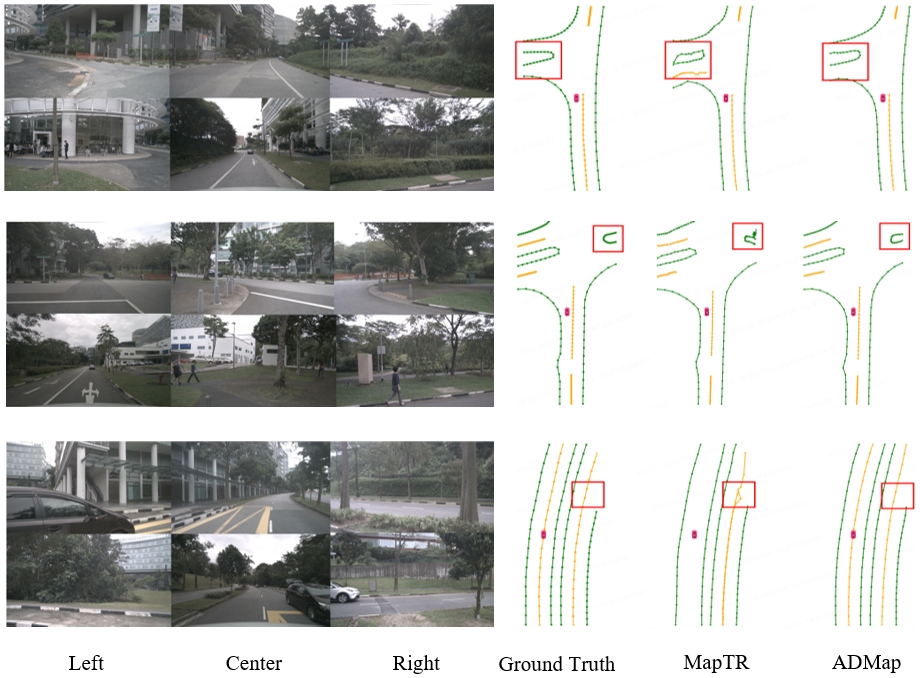}
\end{center}
   \caption{Visualization results of the nuScenes dataset.} % nuScenes 可视化结果
   \label{fig:vis_nus}
\end{figure*}

\begin{figure*}[t]
\small
\begin{center}
\includegraphics[width=\linewidth]{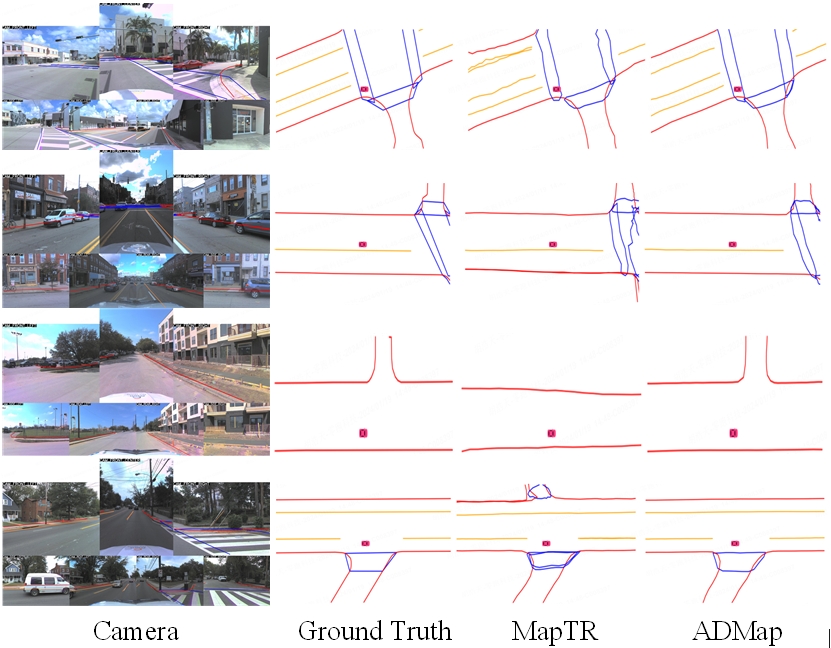}
\end{center}
   \caption{Visualization results of the Argoverse2 dataset} % Argoverse2 可视化结果
   \label{fig:vis_av2}
\end{figure*}